\documentclass[lettersize,journal]{IEEEtran}
\usepackage{amsmath,amsfonts}
\usepackage{algorithm}
\usepackage{array}
\usepackage{textcomp}
\usepackage{stfloats}
\usepackage{url}
\usepackage{verbatim}
\usepackage{graphicx}
\usepackage{cite}
\usepackage{booktabs}
\usepackage{multirow}
\usepackage{color}
\usepackage[table]{xcolor}
\usepackage{nicematrix}
\usepackage{threeparttable}
\usepackage{makecell}
\usepackage{bbding}
\usepackage{ulem}
\usepackage{hhline}
\usepackage{bm}
\usepackage{wrapfig}
\usepackage{enumitem}
\usepackage{caption}
\usepackage{subcaption}

\newcolumntype{L}[1]{>{\raggedright\let\newline\\\arraybackslash\hspace{0pt}}m{#1}}
\newcolumntype{C}[1]{>{\centering\let\newline\\\arraybackslash\hspace{0pt}}m{#1}}
\newcolumntype{R}[1]{>{\raggedleft\let\newline\\\arraybackslash\hspace{0pt}}m{#1}}

\definecolor{myBlue}{HTML}{3ba9db} 
\definecolor{myPink}{HTML}{E7DBCC}
\definecolor{myBlue}{HTML}{5999B6}
\definecolor{myYellow}{HTML}{F1F4AE}

\usepackage{calligra}
\usepackage{algorithmicx}
\usepackage{algpseudocode}

\errorcontextlines\maxdimen

\makeatletter
\newcommand*{\algrule}[1][\algorithmicindent]{\makebox[#1][l]{\hspace*{.5em}\thealgruleextra\vrule height \thealgruleheight depth \thealgruledepth}}%
\newcommand*{\thealgruleextra}{}
\newcommand*{\thealgruleheight}{.75\baselineskip}
\newcommand*{\thealgruledepth}{.25\baselineskip}

\newcount\ALG@printindent@tempcnta
\def\ALG@printindent{%
	\ifnum \theALG@nested>0
	\ifx\ALG@text\ALG@x@notext
	\else
	\unskip
	\addvspace{-1pt}
	\ALG@printindent@tempcnta=1
	\loop
	\algrule[\csname ALG@ind@\the\ALG@printindent@tempcnta\endcsname]%
	\advance \ALG@printindent@tempcnta 1
	\ifnum \ALG@printindent@tempcnta<\numexpr\theALG@nested+1\relax
	\repeat
	\fi
	\fi
}%
\usepackage{etoolbox}
\patchcmd{\ALG@doentity}{\noindent\hskip\ALG@tlm}{\ALG@printindent}{}{\errmessage{failed to patch}}
\makeatother

\newbox\statebox
\newcommand{\myState}[1]{%
	\setbox\statebox=\vbox{#1}%
	\edef\thealgruleheight{\dimexpr \the\ht\statebox+1pt\relax}%
	\edef\thealgruledepth{\dimexpr \the\dp\statebox+1pt\relax}%
	\ifdim\thealgruleheight<.75\baselineskip
	\def\thealgruleheight{\dimexpr .75\baselineskip+1pt\relax}%
	\fi
	\ifdim\thealgruledepth<.25\baselineskip
	\def\thealgruledepth{\dimexpr .25\baselineskip+1pt\relax}%
	\fi
	\State #1%
	\def\thealgruleheight{\dimexpr .75\baselineskip+1pt\relax}%
	\def\thealgruledepth{\dimexpr .25\baselineskip+1pt\relax}%
}

\begin{document}

\title{STAR: Skeletal Token Alignment and Rearrangement for Interaction Recognition}

\author{Yuhang Wen, Mengyuan Liu$^{\dagger}$, Zixuan Tang, Junsong Yuan, Sirui Li, Beichen Ding$^{\dagger}$
\thanks{$^{\dagger}$Corresponding authors: Mengyuan Liu (liumengyuan@pku.edu.cn) and Beichen Ding (dingbch@mail.sysu.edu.cn)}%
\thanks{Yuhang Wen and Zixuan Tang are with School of Intelligent Systems Engineering, Sun Yat-sen University, Shenzhen, China.}%
\thanks{Mengyuan Liu is with State Key Laboratory of General Artificial Intelligence, Peking University, Shenzhen Graduate School, Shenzhen, China.}%
\thanks{Junsong Yuan is with University at Buffalo SUNY, USA.}%
\thanks{Sirui Li is with Department of Computer and Information Science, University of Pennsylvania.}%
\thanks{Beichen Ding is with the School of Advanced Manufacturing \& Southern Marine Science and Engineering Guangdong Laboratory (Zhuhai), Sun Yat-sen University.}%
}

\markboth{Journal of \LaTeX\ Class Files,~Vol.~14, No.~8, August~2021}%
{Wen \MakeLowercase{\textit{et al.}}: STAR: Skeletal Token Alignment and Rearrangement for Interaction Recognition}

\IEEEpubid{0000--0000/00\$00.00~\copyright~2021 IEEE}

\maketitle

\begin{abstract}
Understanding physical human-robot and human-human interactions is a challenging yet emerging topic in 3D vision. While most existing methods rely on skeleton sequences—effective in low-light and privacy-sensitive environments—they face two major challenges: 1) learning and effectively exploiting interaction cues from skeletal data, and 2) compensating for the lack of visual information absent in skeletons alone.
To address these challenges, we propose skeletal token alignment and rearrangement (STAR) for human-robot and human-human interaction recognition.
It learns interaction-specific skeleton features and enriches them using visual cues by aligning skeleton and RGB video representations in a shared latent space.
Specifically, STAR consists of three key components. First, we design a skeleton encoder that captures fine-grained interdependencies using Entity Rearrangement (ER) and Interactive Spatiotemporal Tokens (ISTs). Second, we present Visual Interaction Encoding that introduces a Focus on Interactions (FoI) strategy to attend to spatiotemporal regions relevant to interactions in RGB videos. Finally, these representations are aligned via a contrastive learning objective, with a refinement head further refines predictions.
During training, STAR leverages both skeleton and RGB video data to learn robust, discriminative interaction representations. At inference time, it operates on skeletons alone, retaining visual-informed benefits while preserving skeleton-only efficiency.
Extensive experiments on Chico, HARPER, NTU Mutual 11 and 26 datasets consistently validate our approach by demonstrating superior performance over state-of-the-art methods.
Our code is publicly available at~\url{https://github.com/Necolizer/STAR}.
\end{abstract}


\section{Introduction}

\begin{figure}[t]
    \begin{center}
    \includegraphics[width=0.95\columnwidth]{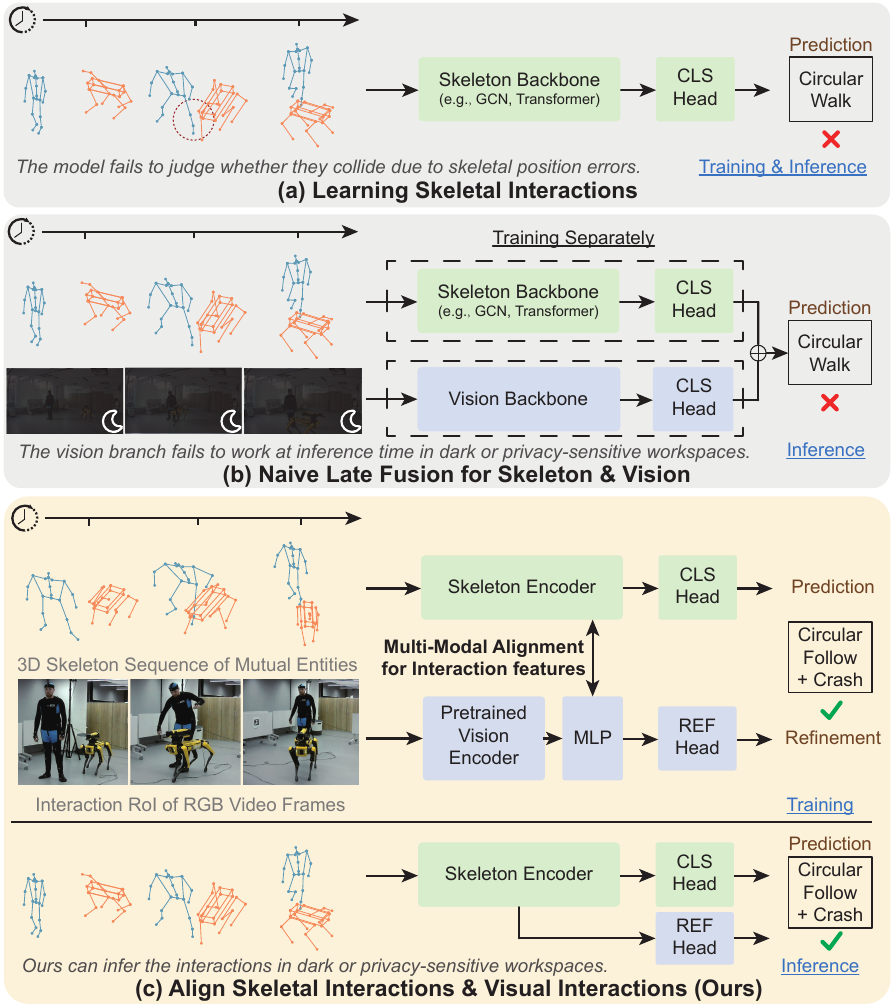}   
    \end{center}
    \caption{Ways to learn interactions. (a) Most existing methods leverage skeleton backbones to model skeletal interactions, which lack visual cues to understand complex interactions. (b) Methods adopting fusion strategies suffer from inefficient inference. Their vision branch cannot work at inference time in dark environments or privacy-sensitive scenarios. (c) Our proposed STAR framework aligns multi-modal interaction features during training to learn more distinctive representations of diverse interactions. It benefits inference with only skeletons, preserving efficiency and efficacy.}
    \label{fig:teaser}
    \vspace{-1.5em}
\end{figure}

\IEEEPARstart{I}{nteraction} recognition task aims to understand mutual dyadic interactions within a scene, such as human interacting with a robot or another human. Examples of such interactions include \textit{robot following human in a circle} (human-robot interactions) and \textit{high-five} (human-human interactions).
Unlike conventional action recognition tasks~\cite{10598383,10287687,10452819,10418536,10321683,9667321,pivit2024CVPR,mmnet2023,10122710,ilic2024selective,Rajasegaran_2023_CVPR,9975251}, this task emphasizes learning spatiotemporal interactions, which is a challenging and essential issue in human-robot interaction (HRI) or collaboration (HRC)~\cite{8438947,jahangard2024jrdbsocial,wen2023interactive,chico2022,HARPER2024}, 2D/3D scene understanding~\cite{10460178,10036100,ren2024spikepoint,jiang2024scaling,Plizzari2024-sp,Wang_2024_CVPR,liang2024intergen,m2i2015}, human-centric models~\cite{Zhu_2023_ICCV,Hong_2022_CVPR,Chen_2023_CVPR,Tang_2023_CVPR,yuan2023hap,Ci_2023_CVPR,wang2023hulk,khirodkar2024_sapiens}, and numerous other areas related to human motion analysis.

\IEEEpubidadjcol

Skeletons have emerged as a prominent modality for modeling motions and interactions~\cite{8636161,Peng2024,10428035,10288273,10.1007/s11263-024-02070-2,Wang_2023_CVPR,Zhou_2023_CVPR,Ng_2020_CVPR}, well representing the interdependent physical dynamics in a compact format. Additionally, skeletal representations offer excellent real-time performance at inference~\cite{FSNET2020}.
However, skeleton-based approaches face two main challenges in interaction recognition. First, prior methods have not fully exploited interaction features in skeletons~\cite{LSTM-IRN2022,igformer2022,GDCN2023,liu2024learning,meformer2024}. Their architectural designs often either lack interaction modeling or are inapplicable to skeleton-based human-robot interactions~\cite{wen2023interactive}. Second, they usually fall short in distinguishing between similar and complex interactions, as shown in Fig.~\ref{fig:teaser} (a), because skeletons lack visual cues necessary for effective disambiguation. Some methods try to solve this problem with multi-modal fusion strategies, but they may fail in low-light or privacy-sensitive scenarios where using RGB videos is impractical.
This leads to a natural question: \uline{How can we better exploit skeletal interaction features and enhance them with visual cues?}

To address the aforementioned problems, we propose a framework based on Skeletal Token Alignment and Rearrangement (STAR) for human-robot and human-human interactions.
Our main insight lies in \uline{leverage permutation-invariance and interactive spatiotemporal local features to more effectively exploit skeletal features, while further enhancing these representations by aligning skeleton-vision feature pairs to achieve more distinctive interactions}.
Specifically, STAR consists of 3 key components. (1) A skeleton encoder. We present Entity Rearrangement, an approach to preserve permutation-invariance among mutual entities and stabilize the optimization of the backbone. Additionally, addressing the limitations of previous architecture designs—such as insufficient interaction modeling and reliance on subject priors—we propose Interactive Spatiotemporal Tokens to represent interactive spatiotemporal local features for interacting skeleton sequence. Building on ISTs, we develop a backbone that effectively encodes skeletal interactions. (2) Visual Interaction Encoding. STAR incorporates Focusing on Interactions to concentrate on the spatiotemporal Region of Interest (RoI) of interactions in RGB videos. (3) A multi-modal interaction alignment framework. We employ a contrastive objective to align RoI features from a pretrained vision encoder with global token features from our skeleton encoder, facilitating disambiguation through visual prompts. Furthermore, we refine the predictions using an auxiliary refinement head. As illustrated in Fig.~\ref{fig:teaser} (c), STAR leverages both skeletons and videos during training, while using only skeletons at inference for efficacy and privacy.

Our main contributions are three-fold:
\begin{itemize}
    \item We propose STAR, a framework for learning human-robot and human-human interactions. STAR is the first to introduce multi-modal interaction alignment into skeleton-based interaction recognition. During training, it aligns interaction features from both skeletons and videos to learn highly discriminative interaction representations, thereby benefiting skeleton-only inference.
    \item Specifically, STAR comprises two encoders and a multi-modal interaction alignment framework that facilitates disambiguation through visual cues. Our skeleton encoder preserves the inherent permutation-invariance in interactions and captures interdependencies between interactive spatiotemporal local features. The visual encoder adopts a Focus on Interactions strategy to target and encoder features related to interactions. The alignment framework further improves the learning of distinguishable representations through contrastive learning in a shared latent space of skeletons and visual RoI.
    \item Extensive experiments on Chico, HARPER, NTU Mutual 26, and NTU Mutual 11 datasets consistently verify STAR, demonstrating its superior performance compared to other state-of-the-art methods.
\end{itemize}

This paper is an extension of our conference paper~\cite{wen2023interactive}. The new contributions of this paper include:
\begin{itemize}
    \item We conduct a more comprehensive analysis of our skeleton encoder, including why Entity Rearrangement works for skeletal interactions and the advantages of Interactive Spatiotemporal Tokens in learning interactive spatiotemporal features. These show STAR's capability to better capture and exploit skeletal interactions.
    \item To further enhance interaction representations, we introduce a novel multi-modal alignment framework that aligns skeleton-vision interaction representations in a shared latent space. This framework consists of three key components: first, we propose Focusing on Interactions to concentrate on interactions in videos both spatially and temporally; second, we adopt a contrastive objective to align positive pairs of skeleton and vision features in the shared latent space; finally, we present a test-time refinement strategy with the refinement head.
    \item We conduct extensive experiments on two new HRI datasets, Chico~\cite{chico2022} and HARPER~\cite{HARPER2024}, demonstrating how our approach aids in supervising physical HRI. Additionally, we compare our method against more state-of-the-art approaches on NTU Mutual 26 and NTU Mutual 11 datasets. Furthermore, we benchmark various state-of-the-art pretrained vision encoders within our framework.
\end{itemize}

The rest of this paper is organized as follows: We first discuss the related work in Section~\ref{sec:relatedwork} and then detail our proposed STAR in Section~\ref{sec:method}. Section~\ref{sec:exp} provides the experimental results. Section~\ref{sec:conclusion} finally concludes this paper.

\section{Related Work}
\label{sec:relatedwork}
\subsection{Skeleton-based Action \& Interaction Recognition}
Supported by the development of large human activity datasets~\cite{NTU60,NTU120,NW-UCLA,liu2017pkummd,Das_2019_ICCV,Ohkawa_2023_CVPR,Yin_2023_CVPR,xu2024inter,Liang2024}, skeleton-based action recognition has emerged as a prominent area of research, particularly in designing effective architectures for recognizing individual actions. Early approaches~\cite{Co-LSTM2016,ST-LSTM2016, GCA2017, VA-LSTM2017, 2s-GCA2018,zhang2019view} tended to capturing long-term contexts with recurrent architectures. Subsequently, methods based on Graph Convolution Network (GCN) became mainstream up till today~\cite{ST-GCN2018,AS-GCN2019, 2s-AGCN2019,MS-G3D2020,CTR-GCN2021,InfoGCN2022,hdgcn2023,9329123,duan2022pyskl,liu2023tsgcnext,degcn2024tip,blockgcn2024CVPR,10598383}.
DeGCN~\cite{degcn2024tip} introduced deformable sampling locations on spatiotemporal graphs, enhancing the perception of discriminative receptive fields.
BlockGCN~\cite{blockgcn2024CVPR} proposed an efficient refinement to graph convolutions to address redundancy issues.
Recent advancements have incorporated self-attention mechanisms into spatiotemporal modeling of skeletons~\cite{dstanet2020,STSA-Net2023,zhou2022hypergraph,8943103,long2023step,MAMP_2023_ICCV,PSUMNet2023,do2024skateformer}.
However, when adapting these models for interaction recognition tasks, most implementations rely on a naive late fusion strategy~\cite{CTR-GCN2021,InfoGCN2022,hdgcn2023,dstanet2020,STSA-Net2023,zhou2022hypergraph}. Unfortunately, they often yield unsatisfactory performance due to insufficient interaction learning~\cite{wen2023interactive}.

To learn interactions based on skeletons, most interactive action recognition methods employ specially designed modules that leverage subject priors~\cite{7577818,6890714,LSTM-IRN2022,igformer2022,GDCN2023,liu2024learning,meformer2024}.
For example, IGFormer~\cite{igformer2022} utilized prior knowledge about human body structure to build up co-attention mechanisms for interaction recognition.
Similarly, me-GCN~\cite{liu2024learning} extracted adjacency matrices for each person and modeled the mutual constraints between them.
Furthermore, some recent works abstract interactions to handle a variety of interaction types~\cite{wen2023interactive,wen2024chase}.
For instance, ISTA-Net~\cite{wen2023interactive} unified the recognition task of interactions across various graph types~\cite{LSTM-IRN2022,igformer2022,GDCN2023,SkeleTR2023,liu2024learning,meformer2024,Assembly101,Wen_2023_CVPR,H2O_TA-GCN2021,H+O2019,shamil2024HandFormer,H2OTR2023CVPR,mucha2024perspective} without relying on adjacency definitions based on subject-type-specific priors.
Moreover, CHASE~\cite{wen2024chase} proposed a normalization methods for individual encoders to benefit recognition in multi-entity settings, including interactions and group activities.
However, these approaches rely solely on skeletons as the modality for modeling mutual interactions, lacking visual prompts to comprehend complex and ambiguous interactions, which limits their performance. To address this issue, our method introduces a multi-modal interaction alignment framework, effectively aligning interaction features from both skeletons and videos within a shared space.

\begin{figure*}[t]
    \begin{center}
    \includegraphics[width=\textwidth]{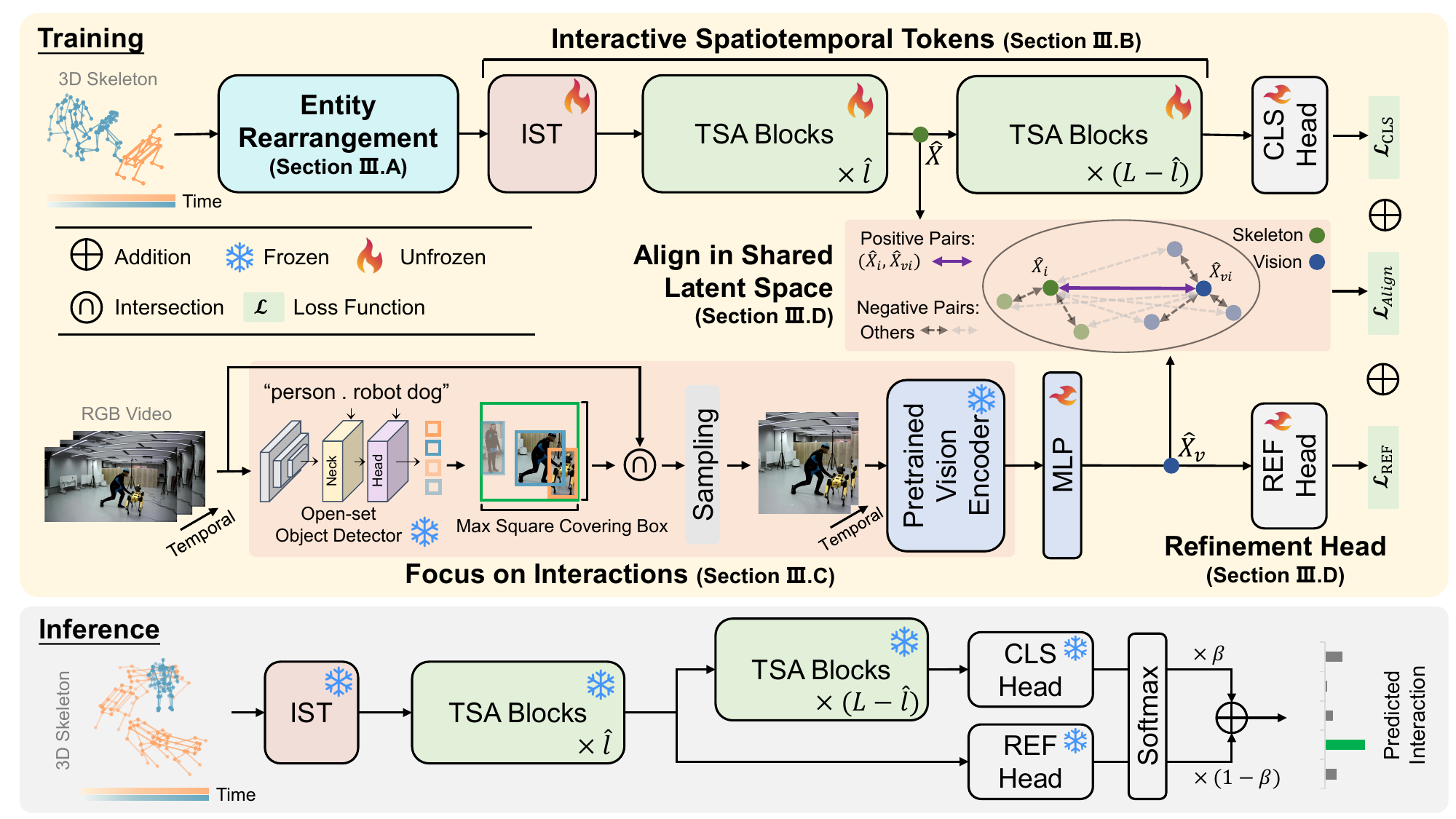}   
    \end{center}
    \caption{The overall framework of the proposed STAR for human-robot and human-human interactions. STAR consists of an encoder to learn skeletal interactions, an encoder to embed visual interactions, and a multi-modal interaction alignment framework to enable disambiguation with visual cues. Our skeleton encoder incorporates Entity Rearrangement to maintain the permutation-invariance of mutual entities and captures interdependencies using Interactive Spatiotemporal Tokens. Visual Interaction Encoding introduces a Focusing on Interactions approach, which targets the spatiotemporal regions of interactions in RGB videos and encodes it with a vision encoder. In our alignment framework, a contrastive objective aligns skeleton-vision feature pairs from the skeleton and vision encoders. Finally, STAR refines inference with an auxiliary refinement head.}
    \label{fig:framework}
    \vspace{-1.2em}
\end{figure*}

\subsection{Multi-modal Learning in Action Recognition}
As multi-modal models become increasingly prevalent in computer vision tasks, recent advancements in skeleton-based action recognition have focused on leveraging diverse modalities to enhance skeletal approaches. There are two main strategies: multi-modality fusion and multi-modality alignment.
Multi-modality fusion incorporates different modalities during both training and evaluation, aiming to fully exploit all available data. Previous studies have made significant progresses in fusing various modalities with skeletal data, including widely adopted RGB videos~\cite{mmnet2023,10235872,Kim_2023_ICCV,vpnpp2022pami,Duan_2022_CVPR,10030830,Shah_2023_WACV}, human parsing maps~\cite{liu2024explore}, motion maps~\cite{dscnet2024}, intra-skeleton modality streams~\cite{degcn2024tip,PSUMNet2023}. This technique appears intuitive and straightforward, using multi-modal data at both training and inference stages. However, it usually increases the complexity and verbosity of the framework, resulting in costly and inefficient inference.
In contrast, multi-modality alignment employs skeletons exclusively during evaluation, aiming to enrich skeletal representations through feature alignment. Several cutting-edge works~\cite{xu2023language,xiang2023gap,liu2024mmcl} have begun to explore this strategy by designing their auxiliary optimization objectives with corresponding RGB patchworks or texts.
IPL~\cite{Wang_2021_ICCV} was proposed for egocentric hand-object interactions in only RGB videos, while our work focuses on multi-modal modeling of human-human and human-robot interactions using 3D skeletons and vision. C\(^2\)VL~\cite{chen2025C2VL} pulled vision-language knowledge prompts and corresponding skeletons closer for human actions in a self-supervised learning paradigm, while our work aligns skeleton-vision pairs in the end-to-end interaction recognition training.
To the best of our knowledge, our method is the first to leverage multi-modal alignment to learn both human-robot and human-human interactions. It benefits from learning with visual cues and preserves the advantages of inferring with sole skeletons.

\section{STAR}
\label{sec:method}

The goal of our task is to classify human-robot and human-human interactions based on their 3D skeleton sequences. This can be expressed as finding the parameterized model \(f_\theta\) with its optimal \(\hat{\theta}\) for the mapping \(f: X\mapsto Y\), where \(X\) represents the interacting skeleton sequences and \(Y\) denotes their corresponding one-hot interaction labels. Let \(E\in \mathbb{Z^+}\) be the number of mutual entities engaged in an interaction over a period of time \(T\in \mathbb{Z^+}\). Each entity may have up to \(J\in \mathbb{Z^+}\) joints, each with \(C\)-dimensional coordinates (\(C=3\) for this task). Zero-padding is applied to entities with fewer than \(J\) joints. Consequently, the input skeleton sequence of an interaction is defined as \(X \in \mathbb{R}^{C\times T\times J \times E}\).

Fig.~\ref{fig:framework} depicts the overview of our proposed STAR for learning human-robot and human-human interactions. It comprises an attention-based encoder to jointly learn the spatial, temporal, and interactive relations of interacting subjects, along with a multi-modal interaction alignment framework to align hidden states across distinct modalities. Specifically, given a 3D skeleton sequence of an interactive action, our end-to-end framework aims to infer the corresponding category of interaction. First, STAR employs Entity Rearrangement during training to preserve the equivalence of unordered subjects. Next, a 3D sliding window tokenizes the skeleton sequences into Interactive Spatiotemporal Tokens, which are then processed by \textit{L}-layer Token Self-Attention Blocks to learn token-level interdependencies. During training, we adopt multi-modal interaction alignment to align skeleton features with vision RoI features. The latter is obtained from a pretrained vision encoder followed by a learnable multi-layer perceptron (MLP). In the inference stage, we ensure efficiency using only concise skeletons. Additionally, with an optimized refinement head, we employ a test-time refinement strategy to achieve refined predictions.

\subsection{Entity Rearrangement}
\label{subsec:er}

\begin{figure}[t]
    \begin{center}
    \includegraphics[width=\columnwidth]{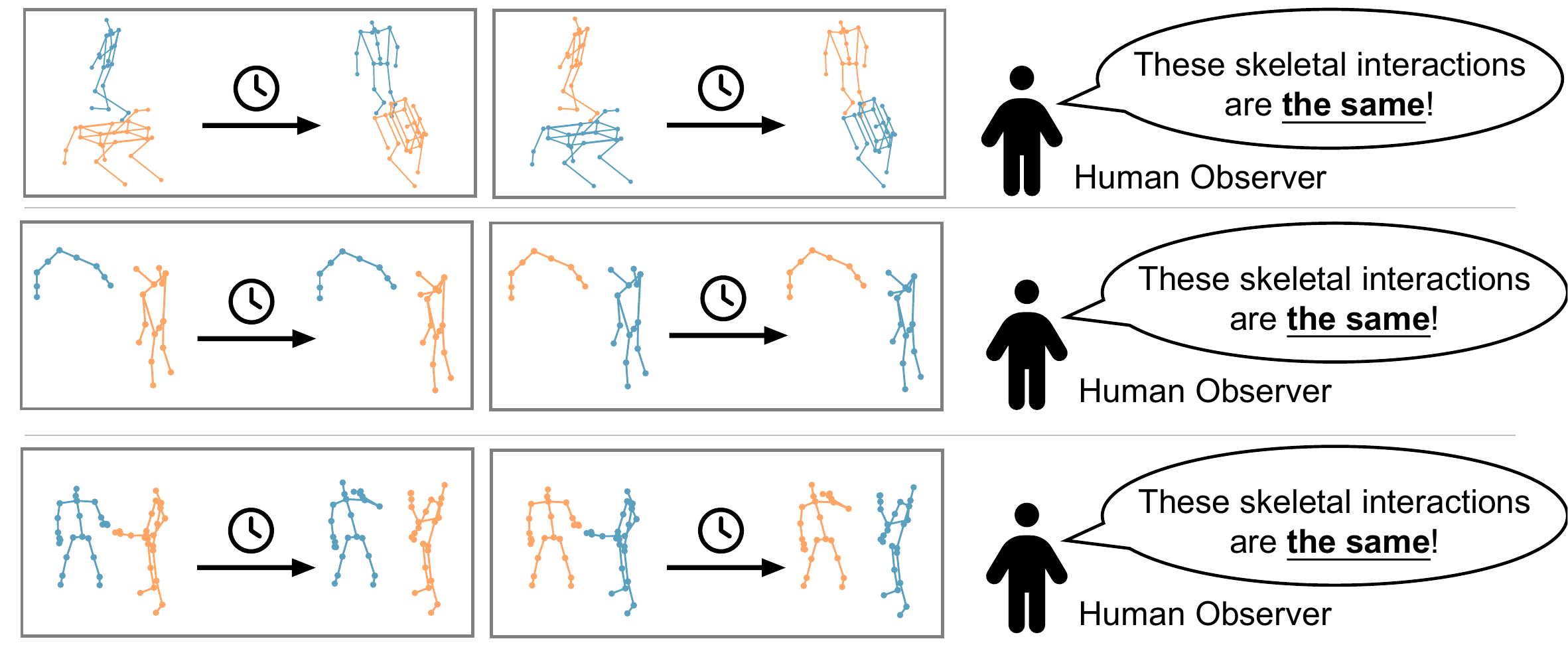}   
    \end{center}
    \caption{Motivation of Entity Rearrangement. We observe that the category of interaction remains unchanged regardless of the order of mutual entities. The entity order in a skeleton sequence is indicated using blue and orange.}
    \label{fig:er}
    \vspace{-1.2em}
\end{figure}

As illustrated in Fig.~\ref{fig:er}, mutual entities in an interaction can be arranged in any order while still representing the same interaction. It implies that the skeletons of unordered entities exhibit permutation-invariance along the entity dimension. This observation inspires a simple yet effective way to eliminate orderliness when modeling interactions, termed Entity Rearrangement (ER).

We begin by formulating ER through a special symmetric group. Consider the finite set \(X_s=\{x_1, \dots, x_E\}\), where \(x_i\in\mathbb{R}^{C\times T\times J}\) is the spatiotemporal keypoints of an entity \(i\). A permutation is defined as a one-to-one mapping of the set onto itself, formulated as
\begin{equation}
    \pi = \begin{pmatrix}x_1&\dots&x_E\\x_{k_1}&\dots&x_{k_E}\end{pmatrix},
\end{equation}
where \(k_1, \dots, k_E\) a specific permutation of the indices \(1, \dots, E\). The set of all permutations on \(X_s\) forms a symmetric group (a permutation group), denoted by \(S_E\).

Consider observations \(O_1, \dots, O_n\in \mathcal{O}\) sampled i.i.d. from a probability distribution \(\mathbb{P}\) over the sample space \(\mathcal{O}\), where each observation of interaction includes a skeleton sequence and its label \(O_i=\{X_{si}, Y_i\}\). Since the category of interaction remains unchanged regardless of the order of the entities, we assume the skeletal data of interactions is permutation-invariant in entity dimension. Consequently, for any group element \(\pi\in S_E\) and almost any \(O\sim\mathbb{P}\), we
have an exact equality in distribution:
\begin{equation}
    \label{eq:exactinvariance}
    O=_d \pi O,
\end{equation}
where \(\pi O=\{\pi X_{s}, Y\}\) and \(=_d\) denotes equality in distribution. This exact invariance means that the probability of a skeleton sequence being \textit{shaking hands} is exactly the same as that of a permuted skeleton sequence. Chen \textit{et al.}~\cite{grouptheoretic2020} proved that when the exact invariance holds, the group action decreases the mean squared error of general estimators and reduces variance in empirical risk minimization (ERM).
In our context of learning skeletal interactions, the symmetric group \(S_E\) can reduces variance in maximum likelihood estimation (a special case of ERM), stabilizing the optimization of the optimal estimator \(\hat{\theta}\). \cite{grouptheoretic2020} also analyzed the case under the approximate invariance \(O\approx_d \pi O\), a weaker assumption than Eq.~\ref{eq:exactinvariance} where \(\approx_d\) denotes approximate equality in distribution. In this case, the symmetric group \(S_E\) can also reduce variance.

In implementation, we randomly choose a \(\pi O\) for any \(\pi \in S_E\) during training. Evidently the cardinality \(|X_s|=E\) implies \(|S_E|=E!\), indicating that each permutation has a probability of \(1/E!\) to be chosen. In practice, \(E\) is typically a small integer in most interaction cases, avoiding non-convergent training.

\begin{figure}[t]
    \begin{center}
    \includegraphics[width=\columnwidth]{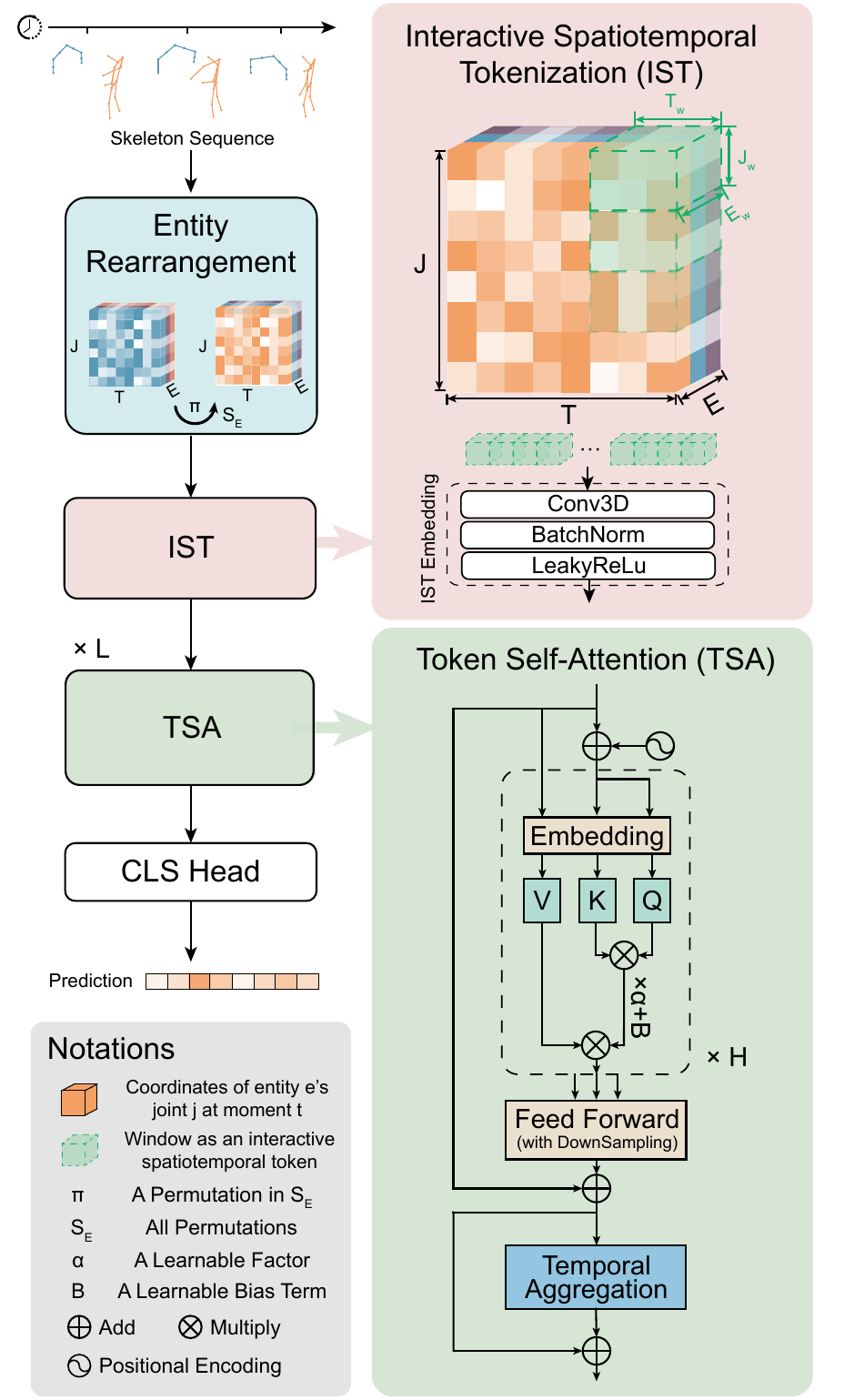}   
    \end{center}
    \caption{The architecture of the skeleton encoder in our proposed STAR. Given a skeleton sequence of an interaction, the skeleton backbone encoders the spatial, temporal, and interactive features simultaneously based on ISTs.}
    \label{fig:net}
    \vspace{-1.0em}
\end{figure}

\subsection{Interactive Spatiotemporal Tokens}
\label{subsec:ist}

Many previous works adopted GCN-based architectures, but they typically require the manual definition of the adjacency matrices~\cite{MS-G3D2020,CTR-GCN2021,InfoGCN2022,hdgcn2023,degcn2024tip,blockgcn2024CVPR}. The prior definition varies depending on the entity type—for example, the physical joint connections of a human body differ significantly from those of a quadruped robot.
This limits their use in most HRI scenarios.
Furthermore, previous models on individual actions attempted to unlock their capability in interaction recognition by a naive late fusion strategy~\cite{CTR-GCN2021,InfoGCN2022,hdgcn2023,dstanet2020,STSA-Net2023}. But this simple walk-around lacks interaction modeling, leading to unsatisfactory performance~\cite{wen2023interactive}.

To address the above issues in previous model design, we introduce the Interactive Spatiotemporal Tokens (ISTs), which represent interactive spatiotemporal local features for interactive skeleton sequences. The overall architecture of our proposed skeleton encoder is illustrated in Fig.~\ref{fig:net}. We delineate the tokenization approach to obtain ISTs and present the Token Self-Attention (TSA) Blocks to capture inter-token correlations. This design simultaneously models the spatial, temporal, and interactive relationships. Moreover, our method does not require priors of the adjacency matrices, offering a unified approach to model both human-robot and human-human interactions.

\textbf{Interactive Spatiotemporal Tokenization.} STAR employs non-overlapping 3D windows to get ISTs. A window is a tuple \(W=(T_{w}, J_{w}, E_{w})\) of integers \(T_{w}, J_{w}, E_{w}\in \mathbb{Z^+}\). It slides along temporal, spatial and entity dimensions, partitioning the input data in a non-overlapping manner. Therefore, the input \(X \in \mathbb{R}^{C\times T\times J \times E}\) is divided into $U = \lceil T/T_{w}\rceil \times \lceil J/J_{w}\rceil \times \lceil E/E_{w}\rceil$ patches in total:
\begin{equation}
\begin{split}
    X_{u_T, u_J, u_E} & = \zeta(X)[:,  u_T T_w : (u_T + 1) T_w, \\
    & u_J J_w : (u_J + 1) J_w, u_E E_w : (u_E + 1) E_w],
\end{split}
\end{equation}
\begin{equation}
    X_{w} = \{ X_{u_T, u_J, u_E}\},
\end{equation}
where \(u_T \in \{0, \dots, \lceil T/T_w \rceil - 1\}\), \(u_J \in \{0, \dots, \lceil J/J_w \rceil - 1\}\), \(u_E \in \{0, \dots, \lceil E/E_w \rceil - 1\}\), \(\zeta\) denotes padding to ensure the shape of \(X\) is dividable by \(W\), and \([:]\) denotes slicing.

The embedding layer for ISTs can be formulated as
\begin{equation}
    X_{ist} = \psi(\sigma (Conv(X_{w}))),
\end{equation}
where \(\psi\) denotes an activation function, \(\sigma\) denotes a normalization layer, and \(X_{ist} \in \mathbb{R}^{C'\times T_{w}\times J_{w}\times E_{w}\times U}\) are the ISTs.

\textbf{Token Self-Attention.} STAR employs self-attention mechanism to capture inter-token relations. Self-attention scores $X^{h}_{l}$ of the $h$-th head in the \(l\)-th TSA Blocks are computed by
\begin{equation}
    X^{h}_{l} = (\alpha\cdot\phi{(\frac{Q^{h}{K^{h}}^{T}}{\sqrt{C^{qkv}_{l}}})} + B)V^{h},
\end{equation}
where \(Q^{h}\), \(K^{h}\), \(V^{h}\) are embedded from the output of the prior TSA Block, \(C^{qkv}_{l}\in \mathbb{Z^+}\) is the number of the attention channels, and \(\phi\) denotes the activation function with a trainable bias term $B \in \mathbb{R}^{U\times U}$ and a trainable balanced factor $\alpha\in \mathbb{R}$\cite{ramapuram2024sigmoidattention}. To avoid over-sparsity in attention maps, we implement multi-head self-attention, where all $X^{h}_{l}$ of $H$ heads in total are concatenated to get $X^{H}_{l}$.

\textbf{Feed Forward Layer with Downsampling.} To enable multi-scale channel capacity, we downsample the features in some certain layers of TSA Blocks. Suppose \(L_D\) is denoted as the set formed by the indices of the downsampling layers, the channel number \(C_{l+1}\) of the output after the \(l\)-th layer is
\begin{equation}
C_{l+1} = \left\{
\begin{aligned}
    2\times C_l, & \mbox{ if }l\in L_D\\
    C_l, & \mbox{ otherwise}
\end{aligned}
\right .,
\end{equation}
where \(l\) indicates the current layer with input channel \(C_l\).

\textbf{Temporal Aggregation.} As temporal aggregation is crucial for modeling actions~\cite{hdgcn2023,wen2023interactive}, STAR adopts a 3D convolution with a temporal kernel size larger than 1 to aggregate inter-frame information in TSA Blocks.

\begin{algorithm}[t]  
	\renewcommand{\algorithmicrequire}{\textbf{Input:}}
	\renewcommand{\algorithmicensure}{\textbf{Output:}}
	\caption{Pseudo Code of STAR}  
	\label{tokenize}
	\begin{algorithmic}[1] 
 	    \Require An input skeleton sequence $X \in \mathbb{R}^{C\times T\times J \times E}$ with its visual representation \(X_{v}\in \mathbb{R}^{C_v}\). A boolean indicator $\eta \in \{0, 1\}$ of training phase. A symmetric group \(S_E\) for the set of all permutations on \(X\) along entity dimension. A 3D sliding window $(T_{w}, J_{w}, E_{w})$, where $T_{w}, J_{w}, E_{w}\in \mathbb{Z^+}$. A number $C'\in \mathbb{Z^+}$ for embedding channel. A negative slope $\gamma\in \mathbb{R}$ for LeakyReLu. \(L\in \mathbb{Z^+}\) layers of TSA Blocks.
		\If{$\eta$}
            \State{$\tilde{X} = \pi X$, for a random \(\pi \in S_E\)}
		\Else
		  \State{$\tilde{X} = \epsilon X$, where \(\epsilon\) is the identity element of \(S_E\)}
		\EndIf
        \State{$X_{w}=$ pad$(\tilde{X})$}
        \State{$X_{w}=$ $X_{w}$.view$(C, T_w, \lceil T / T_w\rceil, J, \lceil J / J_w\rceil, E, \lceil E / E_w\rceil)$}
        \State{$X_{w}=$ Conv3D$_{(1\times 1 \times 1)}(X_{w}, (C, C'))$}
        \State{$X_{1}=X_{ist}=$ LeakyReLu$($BatchNorm3D$(X_{w}), \gamma)$}
        \For{\(1\leq l\leq L\)}
        \State{\(X_{l+1}=\) TSABlock\(_l(X_{l})\)}
        \If{\(l=\hat{l}\) and \(\eta\)}
        \State{\(\mathcal{L}_{align},\hat{X}_{v}=\) Align(mean(\(X_{\hat{l}+1}\)), \(X_{v}\))}
        \EndIf
        \EndFor
        \State{\(\hat{Y}=\) Linear\((X_{L+1})\)}
        \If{\(\eta\)}
            \State{\(\hat{Y}_{v}=\) Linear\(_v(\hat{X}_{v})\)}
            \State{\Return{[\(\hat{Y}\), \(\hat{Y}_{v}\), \(\mathcal{L}_{align}\)]}}
		\Else
		  \State{\(\hat{Y}_{v}=\) Linear\(_v(\)mean\((X_{\hat{l}+1}))\)}
            \State{\Return{[\(\hat{Y}\), \(\hat{Y}_{v}\)]}}
		\EndIf
	\end{algorithmic}
\end{algorithm}

\subsection{Visual Interaction Encoding}
\label{subsec:visual_encode}

To effectively align skeleton and RGB features, it is crucial to first extract meaningful interaction representations from RGB videos. Given that videos often contain task-irrelevant noise, STAR employs a strategy called Focus on Interactions (FoI) to enhance its focus on interaction-relevant regions, both spatially and temporally. Consider an RGB video of \(k\) frames \(V=\{v_i|1\leq i \leq k\}\) recording an interaction. For spatial concentration, STAR employs a pretrained object detector \(\Omega\) to detect the interacting entities in each frame. Open-set detectors~\cite{liu2024groundingdino} can be used here by specifying names of all potential interacting entities in the scene, such as \textit{person} and \textit{robot dog}.
To ensure the images capture not only these entities but also the spatial context between them, STAR utilizes a simple algorithm to find the maximum square bounding box that covers all detected entities in \(V\). As shown in Fig.~\ref{fig:framework}, it gathers all detected bounding boxes of humans and robots across different time steps, represented by blue and orange rectangles, respectively. It then computes a maximum covering box, depicted as a green square. This box then crops each frame, serving as a spatial filter to exclude irrelevant background.
For temporal concentration, STAR samples frames from a trimmed segment of the video, a consecutive sequence of frames obtained by discarding random frames from both the beginning and the end of the raw video. This approach maintains key motions and apex frames crucial to interaction recognition. STAR then encodes the Region of Interest (RoI) of interactions using a pretrained vision encoder \(\Theta\). 
The FoI can be formulated as 
\begin{equation}
    X_{v} = \Theta(\delta(\uplus(\Omega(V))\cap V)),
\end{equation}
where \(\delta\) represents the temporal sampling, \(\uplus\) denotes finding the maximum covering box, and \(\cap\) denotes intersection.

\subsection{Multi-modal Interaction Alignment}
\label{subsec:alignment}
STAR aims to enhance skeletal representations of interactions with multi-modal training data. To this end, we propose a multi-modal interaction alignment framework, which aligns interaction representations within a shared latent space of skeletons and visual RoI features.

\textbf{Align in Shared Latent Space.} To learn distinguishable interaction representations, STAR aligns multi-modal representations of interactions in a shared latent space. After the \(\hat{l}\)-th TSA Block, the intermediate skeleton feature \(\bar{X}_{\hat{l}+1}\in \mathbb{R}^{\hat{C}}\) is hooked for alignment. We denote it as \(\hat{X}\) for simplicity. For vision feature \(X_{v}\in \mathbb{R}^{C_v}\) after FoI, a MLP \(\Psi:\mathbb{R}^{C_v}\mapsto\mathbb{R}^{\hat{C}}\) projects it to a lower-dimensional latent space by reducing its channel dimension, which is consistent with the skeletal hidden state \(\hat{X}\). This space is called a shared latent space. We denote \(\hat{X}_{v}=\Psi(X_{v})\). STAR adopts a contrastive objective based on~\cite{liu2024mmcl,Zhu:2020vf} to align skeleton-vision positive pairs \((\hat{X}, \hat{X}_{v})\). Firstly we define a critic \(\ell(\cdot,\cdot)\) of a given vector pair \((a_i,b_i)\) in a batch of \(N\in \mathbb{Z^+}\) pairs:
\begin{equation}
\label{eq:critic}
    \ell(a_i,b_i)=\frac{d(a_i,b_i)}{\tau} - \log(\sum_{k=1}^{N}e^{\frac{d(a_i,b_k)}{\tau}}+\sum_{\substack{j=1\\(j\neq i)}}^{N}e^{\frac{d(a_i,a_j)}{\tau}}),
\end{equation}
where \(\tau\in \mathbb{R}^{+}\) is the temperature and \(d(\cdot,\cdot)\) is a metric. The first term measures the distance of the positive pair, while the second term weights the logarithmic total distance of all negative pairs. Then the alignment loss is formulated as the averaged score of all pair-wise critics
\begin{equation}
\label{eq:contrastive}
    \mathcal{L}_{Align}=\frac{1}{2N}\sum^{N}_{i=1}[\ell(\hat{X}_i,\hat{X}_{vi})+\ell(\hat{X}_{vi}, \hat{X}_i)],
\end{equation}
where \(N\in \mathbb{Z^+}\) is the batch size.

\textbf{Refinement Head.} We unlock the think-twice ability of STAR by approximating the vision feature \(\hat{X}_{v}\) with the skeleton one \(\hat{X}\) at inference. During training, we train a refinement head \(\xi(\cdot)\) using pretrained vision features \(\hat{X}_{v}\) as input with an auxiliary objective beforehand
\begin{equation}
    \mathcal{L}_{REF} = -\sum_{i}Y_i\log(\mathrm{softmax}(\hat{Y}_{vi})),
\end{equation}
where \(Y_i\) is the ground-truth, and \(\hat{Y}_{vi}\) is the output of the refinement head \(\xi(\cdot)\).

As skeleton-vision positive pairs \((\hat{X}, \hat{X}_{v})\) get aligned during training, STAR then applies skeleton hidden state \(\hat{X}\) to estimate the potential corresponding vision feature \(\hat{X}_{v}\) at inference. The output \(\xi(\hat{X})\) can be used to refine the vanilla classification scores. STAR uses test-time refinement to refine the prediction logits using a convex combination:
\begin{equation}
    \hat{Y}_{logits} = \beta\cdot\mathrm{softmax}(\hat{Y}) + (1-\beta)\cdot\mathrm{softmax}(\xi(\hat{X})),
\end{equation}
where \(\beta\in [0, 1]\) is the coefficient for the convex combination.

\textbf{Training Objective.} To sum up, the complete training loss function can be formulated as
\begin{equation}
    \mathcal{L} = \mathcal{L}_{CLS} + \lambda_1 \mathcal{L}_{Align} + \lambda_2 \mathcal{L}_{REF},
\end{equation}
where \(\lambda_1,\lambda_2 \in \mathbb{R}^{+}\) are trade-off factors, and the first term is the classification loss for the skeleton encoder.

\begin{table*}[t]
	\centering
	\caption{Details of 3D Interaction Recognition Benchmarks}
	\label{table:dataset}
        \begin{threeparttable}
	\begin{tabular}{l|c|c|c|c|c|c|c|c}
		\hline
            \multirow{2}{*}{Benchmarks}&\multicolumn{2}{c|}{3D Skeleton}&\multirow{2}{*}{Egocentric}&\multirow{2}{*}{\#Actions}&\multirow{2}{*}{\#Joints}&\multirow{2}{*}{\#Entities}&\multirow{2}{*}{Data Source}&\multirow{2}{*}{Example(s)}\\	
            \cline{2-3}
            &Human&Robot&&&&&&\\
		\hline
            Chico~\cite{chico2022}&\Checkmark&\Checkmark(Arm)&\XSolidBrush&7&15+9&2&Voxelpose&\textit{surface polishing}\\
            HARPER~\cite{HARPER2024}&\Checkmark&\Checkmark(Quadruped)&\Checkmark(Robot)&15&21+23&2&MoCap&\textit{walk+punch}, \textit{circular walk}\\
            NTU Mutual 11~\cite{NTU60}&\Checkmark&\XSolidBrush&\XSolidBrush&11&25\(\times\)2&2&Kinect v2&\textit{pushing}, \textit{shaking hands}\\
            NTU Mutual 26~\cite{NTU120}&\Checkmark&\XSolidBrush&\XSolidBrush&26&25\(\times\)2&2&Kinect v2&\textit{hit with object}, \textit{follow}\\	
            \hline
	\end{tabular}
       \end{threeparttable}
\end{table*}

\section{Experiments}
\label{sec:exp}
\subsection{Datasets \& Settings}

\textbf{Chico}~\cite{chico2022} is a HRI dataset recording interactions in assembly lines. It provides multi-view videos and 3D skeletons of 20 operators and cobots, engaging in 7 industrial collaborations.

\textbf{HARPER}~\cite{HARPER2024} is the first dataset that depicts physical dyadic interactions between users and quadruped robots. It includes 15 categories of interactions performed by 17 human subjects and a Boston Dynamics Spot, most of which involve physical contact. Notably, Chico~\cite{chico2022} and HARPER~\cite{HARPER2024} are the only two existing HRI datasets that proffer both human and robot 3D skeletons with sufficient interaction categories.

\textbf{NTU Mutual 11}~\cite{NTU60} is a well-recognized benchmark for evaluating human-human interaction recognition methods. It consists of two-person mutual actions from \textbf{NTU RGB+D}~\cite{NTU60} dataset, a large-scale human activity dataset.

\textbf{NTU Mutual 26}~\cite{NTU120}, similarly, is the subset of \textbf{NTU RGB+D 120}~\cite{NTU120} focusing on mutual actions. This dataset features more similar fine-grained interaction categories, e.g, \textit{hitting with object}, \textit{patting on back}, \textit{pushing}, \textit{punching}, etc.

\begin{table*}[t]
        \renewcommand\arraystretch{1.3}
	\centering
	\caption{Comparisons with State-of-the-Art Methods on 3D Interaction Recognition Datasets}
	\label{tab:sota}
        \begin{threeparttable}
	\begin{NiceTabular}{l|c|C{10mm}|C{10mm}|C{10mm}|C{10mm}|C{10mm}|C{10mm}|c}[colortbl-like]
		\hline
		  \multicolumn{1}{c}{\multirow{2}{*}{Methods}}&
            \multirow{2}{*}{Venue}&
            \multicolumn{2}{c|}{NTU Mutual 26 (\%)}&
            \multicolumn{2}{c}{NTU Mutual 11 (\%)}&
            \multicolumn{2}{c|}{Chico (\%)}&
            \multirow{2}{*}{HARPER (\%)}\\
            \cline{3-8}

		  &
            &
            X-Sub&
            X-Set&
            X-Sub&
            X-View&
            IR&
            UCD&
            \\

            \hline

            \multicolumn{9}{c}{Training: Pose\quad Inference: Pose}
            \\

            \hline

            \rowcolor{myYellow!15}
            CTR-GCN\cite{CTR-GCN2021}&
            ICCV'21&
            89.32&
            90.19&
            95.94&
            98.32&
            N/A&
            N/A&
            N/A\\

            \rowcolor{myYellow!15}
            LSTM-IRN\cite{LSTM-IRN2022}&
            TMM'22&
            77.70&
            79.60&
            90.50&
            93.50&
            -&
            -&
            -\\

            \rowcolor{myYellow!15}
            InfoGCN\cite{InfoGCN2022}&
            CVPR'22&
            90.22&
            91.13&
            95.51&
            97.76&
            N/A&
            N/A&
            N/A\\

            \rowcolor{myYellow!15}
            IGFormer\cite{igformer2022}&
            ECCV'22&
            85.40&
            86.50&
            93.60&
            96.50&
            N/A&
            N/A&
            N/A\\

            \rowcolor{myYellow!15}
            STSA-Net\cite{STSA-Net2023}&
            Neuro.'23&
            90.20&
            90.97&
            95.96&
            98.47&
            -&
            -&
            78.71\\

            \rowcolor{myYellow!15}
            HD-GCN\cite{hdgcn2023}&
            ICCV'23&
            88.25&
            90.08&
            95.58&
            97.93&
            N/A&
            N/A&
            N/A\\

            \rowcolor{myYellow!15}
            SkeleTR~\cite{SkeleTR2023}&
            ICCV'23&
            87.80&
            88.30&
            -&
            -&
            -&
            -&
            -\\

            \rowcolor{myYellow!15}
            ISTA-Net~\cite{wen2023interactive}&
            IROS'23&
            90.56&
            91.72&
            -&
            -&
            83.42&
            77.89&
            79.63\\

            \rowcolor{myYellow!15}
            GDCN~\cite{GDCN2023}&
            TPAMI'23&
            85.80&
            92.10&
            -&
            -&
            N/A&
            N/A&
            N/A\\  

            \rowcolor{myYellow!15}
            AHNet-Large~\cite{WANG2024110478}&
            PR'24&
            86.43&
            86.64&
            90.85&
            93.38&
            -&
            -&
            -\\

            \rowcolor{myYellow!15}
            ME-Former~\cite{meformer2024}&
            Biomi.'24&
            90.84&
            91.33&
            95.37&
            97.60&
            N/A&
            N/A&
            N/A\\

            \rowcolor{myYellow!15}
            BlockGCN~\cite{blockgcn2024CVPR}&
            CVPR'24&
            88.40&
            89.84&
            95.05&
            97.11&
            N/A&
            N/A&
            N/A\\

            \rowcolor{myYellow!15}
            DeGCN~\cite{degcn2024tip}&
            TIP'24&
            90.84&
            90.36&
            95.94&
            97.95&
            N/A&
            N/A&
            N/A\\

            \rowcolor{myYellow!15}
            CHASE+CTR-GCN~\cite{wen2024chase}&
            NeurIPS'24&
            91.30&
            92.34&
            96.45&
            98.83&
            N/A&
            N/A&
            N/A\\

            \rowcolor{myYellow!15}
            me-GCN~\cite{liu2024learning}&
            THMS'25&
            90.00&
            90.00&
            95.50&
            98.20&
            N/A&
            N/A&
            N/A\\

            \hline

            \multicolumn{9}{c}{Training: Multi-Modality\quad Inference: Pose}
            \\

            \hline

            \rowcolor{myBlue!10}
            GAP~\cite{xiang2023gap}&
            ICCV'23&
            89.27&
            90.60&
            -&
            -&
            N/A&
            N/A&
            N/A\\

            \rowcolor{myBlue!10}
            MMCL~\cite{liu2024mmcl}&
            ACMMM'24&
            91.42&
            91.66&
            96.43&
            98.55&
            N/A&
            N/A&
            N/A\\

            \rowcolor{myBlue!10}
            \textbf{STAR (Ours)}&
            -&
            \textbf{92.23}&
            \textbf{92.75}&
            \textbf{96.63}&
            \textbf{99.05}&
            \textbf{84.63}&
            \textbf{78.50}&
            \textbf{81.07}\\
		
		\hline
	\end{NiceTabular}
        \begin{tablenotes}
         \item N/A: Not applicable to heterogeneous interacting entities in human-robot interactions.
        \end{tablenotes}
        \end{threeparttable}
        \vspace{-1.0em}
\end{table*}

\textbf{Settings.} Details of the above datasets are listed in Table~\ref{table:dataset}. In our experiments, we use 3D pose sequences in joint modality (without intra-skeleton modality fusion, to ensure fair comparisons~\cite{igformer2022,wen2023interactive}) and RGB videos (for alignment) in each dataset.
In Chico, we follow the train/validation/test split in~\cite{chico2022}.
We pick continuous 300-frame clips with 50-frame stride in long skeleton sequences as HRI samples.
Two evaluation protocols are adopted: \textit{Interaction Recognition} (IR) is to recognize different types of HRI, and \textit{Unexpected Collision Detection} (UCD) is to judge whether an HRI involves at least an unexpected physical collision.
In HARPER, we follow the train/test split in~\cite{HARPER2024} using 30Hz pose sequences.
In NTU Mutual 11, we leverage the Cross-subject (X-Sub) and Cross-view (X-View) settings~\cite{NTU60} for evaluation, while in NTU Mutual 26, we adopt the Cross-subject (X-Sub) and Cross-set (X-Set) protocol~\cite{NTU120}. 
The top-1 accuracy is adopted as the evaluation metric. Some experimental results for comparisons are from related works~\cite{igformer2022, wen2023interactive, meformer2024, wen2024chase}.

\subsection{Implementation Details}
Our experiments are conducted on 8 GeForce RTX 3080Ti GPUs. Grounding Dino~\cite{liu2024groundingdino} is employed as the default pretrained object detector \(\Omega\). A more efficient close-set detector YOLOv5~\cite{yolov5} is employed for human-human interactions to accelerate processing. The default pretrained vision encoder \(\Theta\) is UniformerV2~\cite{UniFormerV2_2023_ICCV}. 
We use mixup~\cite{zhang2018mixup} and manifold mixup~\cite{manimixup2019} strategies during training. In skeleton encoder, we adopt \(L=8\), \(L_D=\{3,5\}\), \(W=(20, 1, 2)\), and \(H=3\). In multi-modal alignment, \(\mathcal{L}_{CLS}\) is implemented by the cross entropy with label smoothing factor 0.05. We set default \(\hat{l}=6\), \(\lambda_1=0.4\), \(\lambda_2=0.7\), and \(\beta=0.8\). \(d(\cdot,\cdot)\) is implemented by cosine similarity. SGD optimizer is used with Nesterov momentum of 0.9, a initial learning rate of 0.1 and a decay rate 0.1 at 80th, 110th, 130th, 140th epoch. Batch size is 64. With the first 5 warm-up epochs, each training process is terminated after 160 epochs. As some hyper-parameters may vary across different benchmarks, please refer to the benchmark-specific configurations in our code repository. All experiments require the type of actors to be specified in advance, as we focus on the Interaction Recognition task, following the widely accepted setting in recent works~\cite{LSTM-IRN2022,GDCN2023,wen2024chase,liu2024learning}. This also ensures a fair comparison with existing approaches.

\subsection{Quantitative Results}

Table~\ref{tab:sota} presents the quantitative comparison of STAR with existing methods on 3D human-robot and human-human interaction recognition datasets. We categorize the baselines into (1) skeleton-based action recognition methods using a late fusion strategy~\cite{CTR-GCN2021,InfoGCN2022,STSA-Net2023,hdgcn2023,blockgcn2024CVPR,degcn2024tip}, (2) skeleton-based interaction recognition methods~\cite{LSTM-IRN2022,igformer2022,SkeleTR2023,wen2023interactive,GDCN2023,WANG2024110478,meformer2024,liu2024learning,wen2024chase}, and (3) action recognition methods leveraging multi-modal alignment~\cite{xiang2023gap,liu2024mmcl}. Note that \uline{most related methods cannot apply to heterogeneous entities in HRI datasets} (Chico and HARPER) without modifying their architectures or defined adjacency matrices. Our proposed STAR is the first to introduce multi-modal interaction alignment in skeleton-based interaction recognition. It outperforms these state-of-the-art methods across all benchmarks by a noticeable margin. Our method outperforms ISTA-Net~\cite{wen2023interactive}, DeGCN~\cite{degcn2024tip}, CHASE~\cite{wen2024chase} with CTR-GCN~\cite{CTR-GCN2021} backbone, and other methods training with only skeletons, highlighting the effectiveness to introduce multi-modal representations into modeling interactions. Specifically, ISTA-Net~\cite{wen2023interactive} and related methods are uni-modal, which fails to classify complicated interactions. STAR’s ability to align skeleton and vision RoI features mitigates this limitation, leading to improved performance. Additionally, we compare our method with action recognition methods using multi-modal alignment. GAP~\cite{xiang2023gap} leverages category-shared text descriptions in training. Similarly, MMCL~\cite{liu2024mmcl} adopts vision-language pairs with manual written queries about the actions. Compared with these approaches, STAR achieves better interaction recognition performance by aligning skeleton and vision RoI features of spatiotemporal interactions. In NTU Mutual 26, STAR outperforms MMCL by 0.81\% and 1.09\% on X-Sub and X-Set criteria, respectively. This is attributed not only to our effective skeleton encoder design, but also to our learning strategy of distinguishable representations, using contrastive learning in a shared latent space of skeletons and visual interaction RoI.

\begin{figure*}[t]
    \begin{center}
    \includegraphics[width=0.9\textwidth]{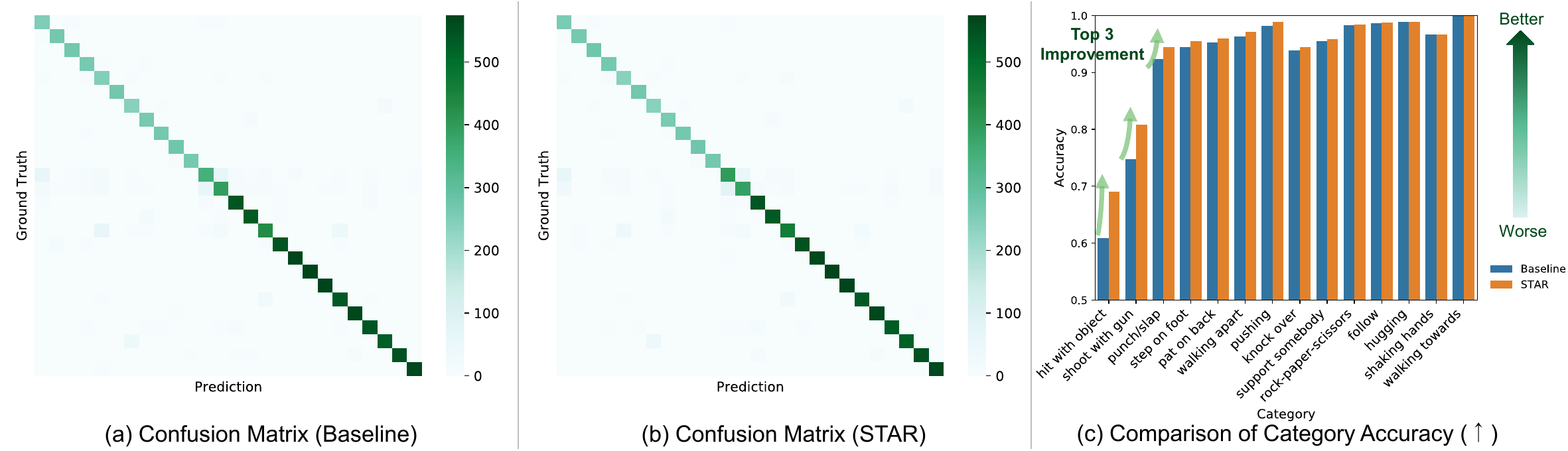}   
    \end{center}
    \caption{Confusion matrices of the baseline (a) and our proposed STAR (b). Their detailed category results (c) demonstrate that STAR effectively recognizes similar, fine-grained skeletal interactions by aligning with visual RoI of interactions, e.g., \textit{hit with object}, \textit{punch/slap}, and \textit{pat on back}.}
    \label{viz:matrix}
    \vspace{-1.0em}
\end{figure*}

\begin{figure}[t]
    \begin{center}
    \includegraphics[width=0.95\columnwidth]{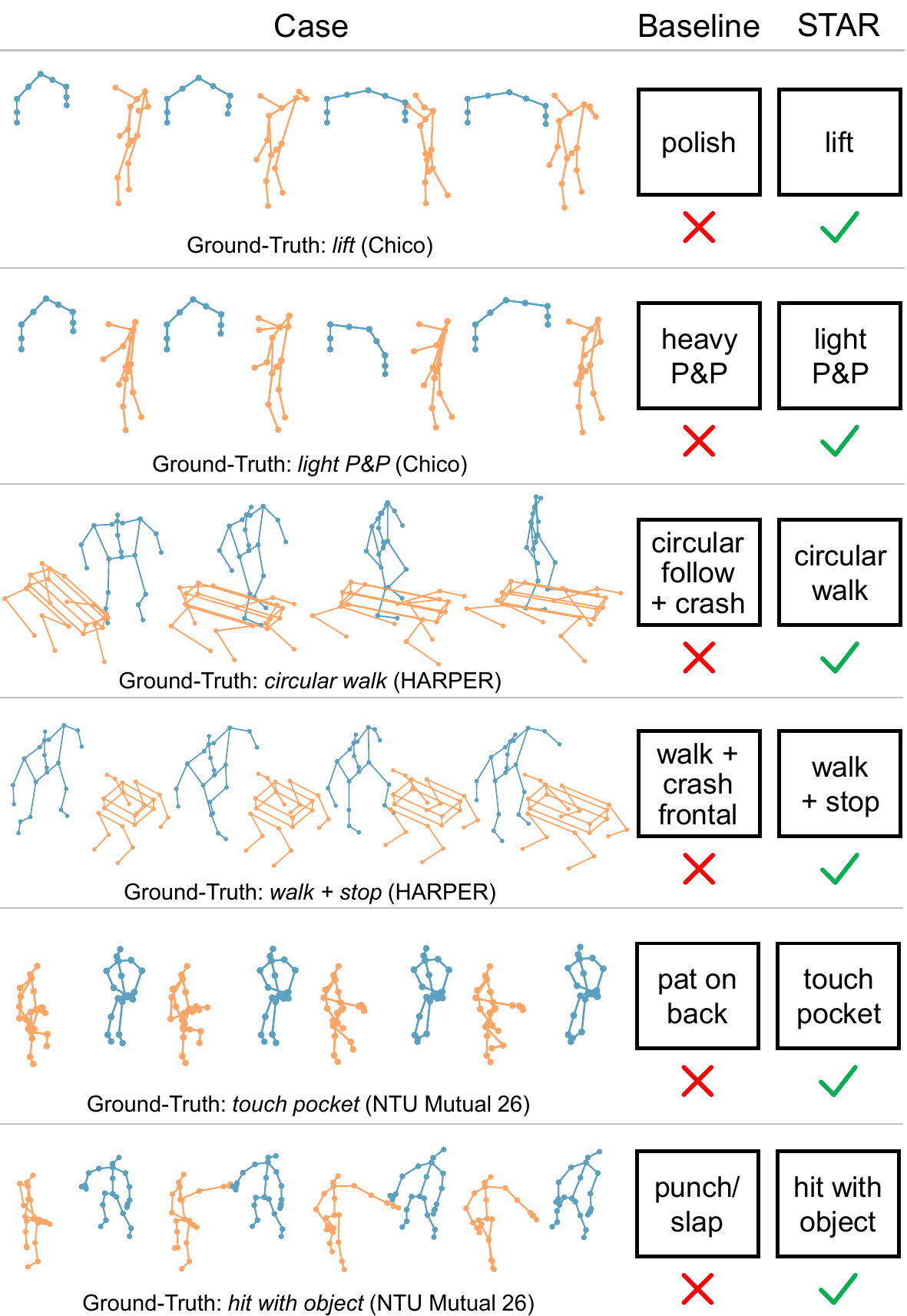}   
    \end{center}
    \caption{Case visualization and analysis of difficult interaction samples. We present confusing interaction samples that make the baseline classifier produce wrong answers. They require visual cues of manipulated objects and contacting parts to be distinguished from other similar interactions.}
    \label{fig:case}
    \vspace{-1.0em}
\end{figure}

\subsection{Qualitative Results}

Fig.~\ref{viz:matrix} shows the confusion matrices and a comparison of category-level scores between the baseline~\cite{wen2023interactive} and our proposed STAR. Compared with the baseline, STAR can achieve better category-level recognition performance, e.g., \textit{hit with object} (+8.17\%), \textit{shoot with gun} (+6.09\%), and \textit{punch/slap} (+2.19\%). It implies training with visual cues of interactions benefits better disambiguation. For example, this alignment framework learns from additional information such as manipulated objects to better recognize \textit{hit with object} and \textit{shoot with gun}. It also learns from exact contacting body parts in visual appearances to differentiate between similar interactions like \textit{punch/slap}, \textit{pushing}, \textit{pat on back}, and \textit{support somebody}. We note that object-related interaction categories (e.g., hit with object) show relatively lower accuracy, mainly due to the dataset's lack of 3D keypoints for manipulated objects. This limits the contextual cues available for skeleton-only inference. We believe that incorporating 3D object keypoints would further improve our method's performance on such categories by enabling finer-grained interaction understanding.

Additionally, Fig.~\ref{fig:case} conducts a case study of hard interaction samples. These interactions involve manipulating objects or close contact, which are confusing for the baseline method. In contrast, our proposed STAR aligns skeletal and visual representations of interactions within a shared latent space during training, facilitating the development of distinct representations. It helps the skeleton encoder to accurately classify these difficult samples at inference stage.

\begin{figure*}[t]
    \begin{center}
    \includegraphics[width=0.9\textwidth]{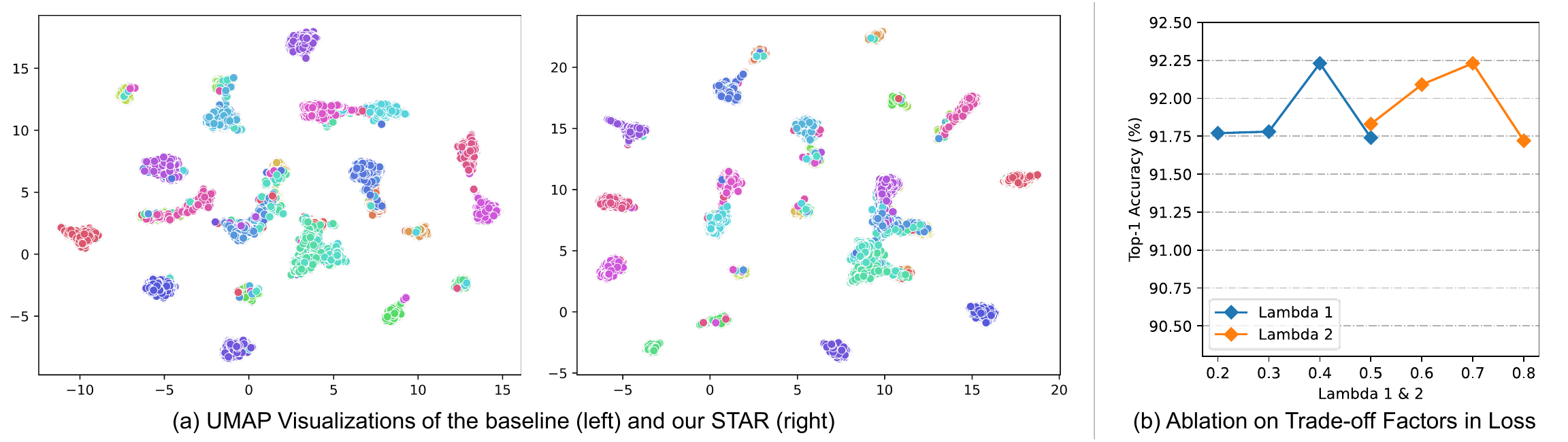}   
    \end{center}
    \caption{(a) UMAP~\cite{mcinnes2018umap-software} visualizations of skeletal representations of interactions on the test set. Compared to the baseline (left), STAR (right) produces more compact and well-separated clusters, indicating that it learns more discriminative interaction representations. This improvement is attributed to the alignment between the skeleton hidden state and visual RoI features during training, which helps STAR better differentiate similar interaction patterns. (b) We conduct ablation on different values of the trade-off factors \(\lambda_1,\lambda_2\) in our training objective.}
    \label{viz:umap}
\end{figure*}

Moreover, the UMAP~\cite{mcinnes2018umap-software} visualization in Fig.~\ref{viz:umap} (a) demonstrates STAR effectively learns more discriminative skeletal representations of interactions compared to the baseline. Each point in the figure corresponds to a test sample's latent representation, color-coded by interaction class. While the baseline exhibits overlapping clusters, STAR produces clearer and more separated ones. This is attributed to our training-time alignment between skeletons and visual RoI features, which enhances the model’s ability to distinguish similar interactions using skeleton data alone at inference.

\subsection{Ablation Study}

\begin{table*}[t]
    \caption{Ablation on key components and hyper-parameters of STAR}
    \label{ablation:key}
    \begin{subtable}[t]{0.42\textwidth}
        \centering
	\begin{tabular}{l|c|c}
            \hline
               Vision Encoder & Input Shape & Acc (\%)\\
		  \hline
                No Alignment & - & 91.50 \\
            \hline
                DINOv2-B~\cite{oquab2023dinov2} & \(1\times 224\times 224\) & 91.52 \\
                Sapiens-0.3B~\cite{khirodkar2024_sapiens} & \(1\times 1024\times 1024\) & 91.54 \\
            \hline
                 VideoMAEv2-B~\cite{wang2023videomaev2}& \(16\times 224\times 224\) & 91.94 \\
                \textbf{UniformerV2-B}~\cite{UniFormerV2_2023_ICCV} & \(8\times 224\times 224\) & \textbf{92.23} \\
            \hline
	\end{tabular}
        \caption{Pretrained Vision Encoders}
	\label{ablation:visionencoder}
     \end{subtable}
    \begin{subtable}[t]{0.28\textwidth}
        \centering
	\begin{tabular}{l|c}
            \hline
                Method & Acc (\%)\\
		  \hline
                \textbf{STAR} & \textbf{92.23} \\
            \hline
                w/o FoI (Resize + Encode) & 91.48\(^{{\color{blue} \downarrow 0.75}}\) \\
                w/o Multi-modal Alignment & 91.50\(^{{\color{blue} \downarrow 0.73}}\) \\
                w/o Refinement Head & 91.72\(^{{\color{blue} \downarrow 0.51}}\) \\
                w/o Test-Time Refinement & 92.18\(^{{\color{blue} \downarrow 0.05}}\) \\
            \hline
	\end{tabular}
        \caption{Multi-modal Interaction Alignment}
	\label{ablation:align}
     \end{subtable} %
    \begin{subtable}[t]{0.3\textwidth}
        \centering
	\begin{tabular}{l|c}
            \hline
                Method & Acc (\%) \\
		  \hline
                \textbf{ISTs} & \textbf{92.23} \\
                Co-attention & 91.76\(^{{\color{blue} \downarrow 0.47}}\)\\
                Late Fusion & 90.62\(^{{\color{blue} \downarrow 1.61}}\)\\
                Channel Concat & 89.21\(^{{\color{blue} \downarrow 3.02}}\)\\
            \hline
                \textbf{w/ ER} & \textbf{92.23} \\
                w/o ER &  91.50\(^{{\color{blue} \downarrow 0.73}}\)\\
            \hline
	\end{tabular}
        \caption{Skeletal Interaction Modeling}
	\label{ablation:ist}
    \end{subtable} %
    \begin{subtable}[t]{0.20\textwidth}
        \centering
	\begin{tabular}{l|c}
            \hline
               Loss & Acc (\%)\\
		  \hline
                \textbf{Contrastive} & \textbf{92.23}\\
                MSE & 91.76\(^{{\color{blue} \downarrow 0.47}}\) \\
                MMD & 91.65\(^{{\color{blue} \downarrow 0.58}}\) \\
            \hline
	\end{tabular}
        \caption{Alignment Loss}
	\label{ablation:loss}
     \end{subtable} %
     \begin{subtable}[t]{0.27\textwidth}
        \centering
	\begin{tabular}{l|c|c}
            \hline
               Layer & Param. & Acc (\%)\\
		  \hline
                10 & 8.19M & 92.11\(^{{\color{blue} \downarrow 0.12}}\)\\
                \textbf{8} & 6.63M & \textbf{92.23}\\
                6 & 5.06M & 91.60\(^{{\color{blue} \downarrow 0.63}}\)\\
                4 & 3.50M & 91.39\(^{{\color{blue} \downarrow 0.84}}\)\\
            \hline
                No Alignment & 6.22M & 91.50\(^{{\color{blue} \downarrow 0.73}}\)\\
            \hline
	\end{tabular}
        \caption{Encoder Layers}
	\label{ablation:layers}
     \end{subtable} %
    \begin{subtable}[t]{0.14\textwidth}
        \centering
	\begin{tabular}{l|c}
            \hline
               \(\hat{l}\) & Acc (\%)\\
		  \hline
                8 & 91.81\(^{{\color{blue} \downarrow 0.42}}\)\\
                7 & 91.93\(^{{\color{blue} \downarrow 0.30}}\)\\
                \textbf{6} & \textbf{92.23}\\
                5 & 92.00\(^{{\color{blue} \downarrow 0.23}}\) \\
            \hline
	\end{tabular}
        \caption{Aligning Block}
	\label{ablation:blk}
     \end{subtable} %
    \begin{subtable}[t]{0.17\textwidth}
        \centering
	\begin{tabular}{l|c}
            \hline
               Shape & Acc (\%)\\
		  \hline
                \textbf{(20, 1, 2)} & \textbf{92.23} \\
            \hline
                (30, 1, 2) & 91.52\(^{{\color{blue} \downarrow 0.71}}\)\\
                (10, 1, 2) & 90.99\(^{{\color{blue} \downarrow 1.24}}\)\\
                (20, 2, 2) & 91.65\(^{{\color{blue} \downarrow 0.58}}\)\\
                (20, 1, 1) & 91.46\(^{{\color{blue} \downarrow 0.77}}\)\\
            \hline
	\end{tabular}
        \caption{Sliding Window Shapes}
	\label{ablation:window}
     \end{subtable} %
     \begin{subtable}[t]{0.19\textwidth}
        \centering
	\begin{tabular}{l|c}
            \hline
               + Noise & Acc (\%) \\
		  \hline
                Baseline & 90.06\(^{{\color{blue} \downarrow 1.05}}\) \\
                \textbf{STAR} & \textbf{91.11} \\
            \hline
               + Mask & Acc (\%) \\
		  \hline
                Baseline & 91.04\(^{{\color{blue} \downarrow 0.97}}\) \\
                \textbf{STAR} & \textbf{92.01} \\
            \hline
	\end{tabular}
        \caption{Test-time Robustness}
	\label{ablation:robustness}
     \end{subtable} %
     \vspace{-2.0em}
\end{table*}

\textbf{Pretrained Vision Encoders.} We benchmark different state-of-the-art pretrained vision encoders in our STAR framework, including image encoders (DINOv2~\cite{oquab2023dinov2} and Sapiens~\cite{khirodkar2024_sapiens}) and video encoders (VideoMAEv2~\cite{wang2023videomaev2} and UniformerV2~\cite{UniFormerV2_2023_ICCV}). As image encoders encode a single image instead of a multi-frame video, we test two common adaptations to video inputs. 1) For DINOv2~\cite{oquab2023dinov2}, we use the frame-averaged feature \(X_{v}\) for alignment. 2) For Sapiens~\cite{khirodkar2024_sapiens}, due to its high-resolution input requirement, we implement frame concatenation (a common practice for video action recognition task~\cite{mmnet2023,liu2024explore,liu2024mmcl}) to construct the input image. Table~\ref{ablation:key} (\subref{ablation:visionencoder}) suggests that video encoders outperform image encoders with the above two adaptation strategies. This interesting observation implies that \uline{complex temporal modeling ability of pretrained video encoders offers more semantically meaningful and discriminative representations}. 

\textbf{Multi-modal Interaction Alignment.} Table~\ref{ablation:key} (\subref{ablation:align}) ablates the key components in our multi-modal interaction alignment framework. First, we replace FoI with a simple pipeline that first resizes the frames and then encodes them with vision encoders. The significant drop in accuracy highlights the importance of concentrating on the RoI of interactions in the videos. Second, we observe a substantial decline in performance when removing the alignment loss, which means STAR without multi-modal alignment. \uline{This result validates the effectiveness of our multi-modal interaction alignment framework.} It also proves that it's effective and necessary to align skeletons and RGB videos, as they provide distinct yet complementary representations of interactions from different sources. Additionally, removing the refinement head during training results in a decrease in accuracy, indicating that STAR benefits from the visual refinement head through disambiguation of interactions in the shared latent space. The result also shows that STAR could recognize interactions better by leveraging our test-time refinement strategy.

\textbf{Skeletal Interaction Modeling.} We compare different approaches to model skeletal interactive relations of spatiotemporal keypoints. 1) \textit{Co-attention} employs weight-shared dual-branch attention blocks for two entities~\cite{igformer2022,liu2024learning}. One branch uses entity features from another branch to query the key-value pairs of its own. 2) \textit{Late Fusion} is a common practice for single-entity action recognition models when adapting to interactive skeletons~\cite{CTR-GCN2021,InfoGCN2022,hdgcn2023,dstanet2020,STSA-Net2023,zhou2022hypergraph}. This strategy means that skeletal interactions are modeled only in classification heads. 3) \textit{Channel Concat} directly concatenates entity skeletons along coordinate dimension. 4) Our proposed \textit{ISTs} fuse interactive features during early tokenization. Table~\ref{ablation:key} (\subref{ablation:ist}) indicates that \textit{ISTs} outperforms the alternatives by 0.47\%, 1.61\% and 3.02\%, respectively, \uline{highlighting the superiority of \textit{ISTs} to model interactive spatiotemporal features.}
Additionally, we validate the effectiveness of Entity Rearrangement by removing this step. As reported in Table~\ref{ablation:key} (\subref{ablation:ist}), the accuracy score declines by 0.73\% without ER, showing that \uline{ER is beneficial for preserving permutation-invariance and stabilizing backbone optimization of the backbone}.

\textbf{Alignment Loss.} Table~\ref{ablation:key} (\subref{ablation:loss}) presents an ablation study of different alignment loss functions, comparing the contrastive objective with Mean Square Error (MSE) and Maximum Mean Discrepancy (MMD). The latter two losses have been used in several recent recognition methods requiring feature alignment~\cite{pivit2024CVPR,InfoGCN2022,wen2024chase}. The contrastive objective, as shown in Eq.~\ref{eq:contrastive}, outperforms both MSE and MMD, enhancing the skeleton encoder by leveraging contrastive learning between intermediate skeleton features and vision-based ones.

\textbf{Number of Skeleton Encoder Layers \& Aligning Blocks.} In Table~\ref{ablation:key} (\subref{ablation:layers}), we scale the skeleton encoder by adding or removing TSA Blocks to evaluate the impact on performance. For each configuration, a TSA Block is added (or removed) at both the top and bottom of the encoder.
Results indicate that 8 encoder layers achieve the best recognition accuracy, as more layers can lead to overfitting. Additionally, Table~\ref{ablation:key} (\subref{ablation:layers}) shows that our alignment strategy, specifically using a MLP \(\Psi\) and a refinement head \(\xi(\cdot)\), only increase the parameters by 0.41M, highlighting the efficiency.
Moreover, we evaluate the performance when aligning skeleton-vision pairs after different \(\hat{l}\)-th TSA Blocks. As reported in Table~\ref{ablation:key} (\subref{ablation:blk}), aligning the intermediate skeletal feature after the 6th block yields the best performance with the benefit of post-alignment modeling.

\textbf{Different Window Sizes \& Trade-off Factors.} We evaluate the performance under various window sizes and balance factors. Table~\ref{ablation:key} (\subref{ablation:window}) demonstrates that given a fixed number of input frames, the window size (20,1,2) leads to the optimal result. Results in Fig.~\ref{viz:umap} (b) suggest the optimal trade-off factors used in our loss function as the default setting.

\textbf{Robustness of STAR.} To evaluate the robustness of STAR, we intentionally corrupt the skeleton sequences with noise and masking at test time. This aims at resembling possible tracking or estimation errors during the inference phase. The noise \(X_n\sim \mathcal{N}(\mu,\,\sigma^{2})\) applied in this experiment is normally distributed with mean \(\mu=0\) and standard deviation \(\sigma=0.01\). Masking strategy is randomly masking the skeleton sequences with probability \(p_m=0.01\). As shown in Table~\ref{ablation:key} (\subref{ablation:robustness}), compared to STAR without multi-modal alignment as baseline, \uline{our STAR can recognize corrupted test-time skeletons more effectively, indicating its better robustness brought by learning distinguishable representations of interactions}.

\begin{figure}[t]
    \begin{center}
    \includegraphics[width=\columnwidth]{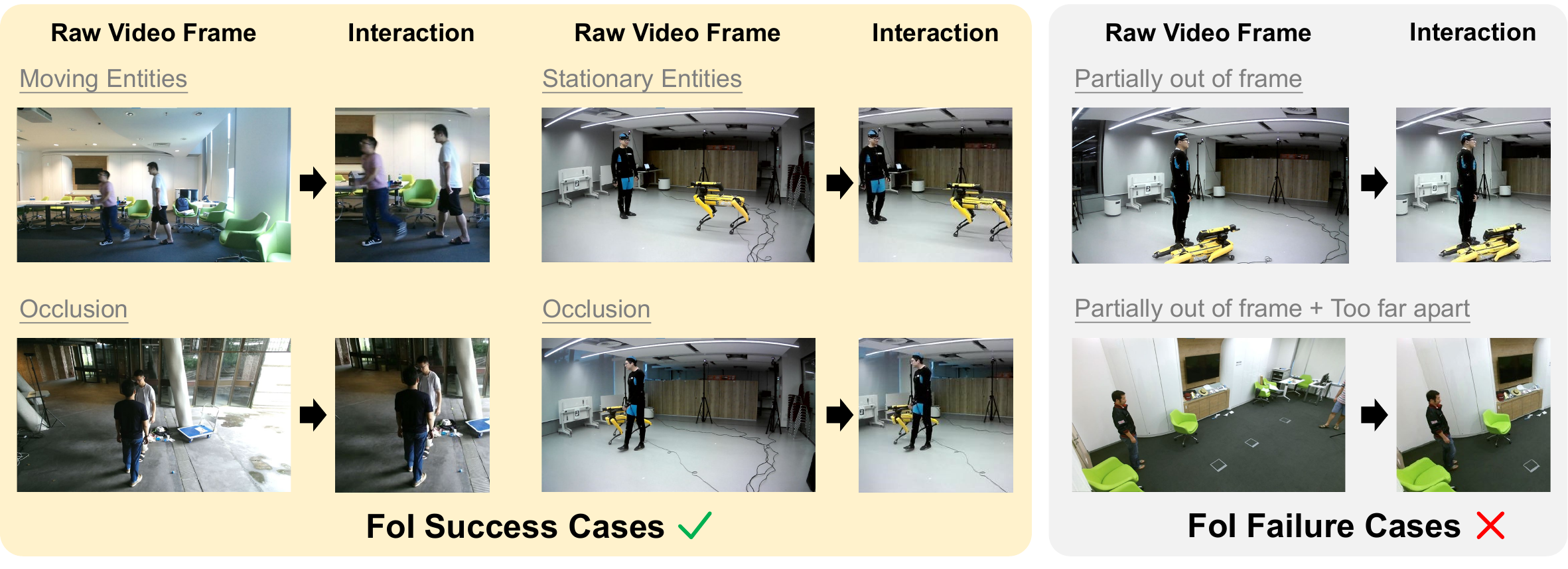}   
    \end{center}
    \vspace{-0.5em}
    \caption{Success and failure cases in FoI. FoI succeeds in both dynamic and static interactions, and remains robust under occlusion. Failure cases are rare and typically occur when entities are partially out of frame or too far apart.}
    \label{fig:sf}
    \vspace{-1.2em}
\end{figure}

\textbf{Success \& Failure Cases of FoI.} To better understand the robustness of FoI, we analyze typical success and failure cases in Fig.~\ref{fig:sf}. FoI successfully captures interaction regions in a wide range of settings, including both dynamic and static entities, as well as moderate occlusion. We note that failure cases are rare, and usually involve situations where interacting entities are partially out of frame or positioned far apart. Despite their infrequency, such cases suggest possible gains in future extensions.

\section{Conclusion}
\label{sec:conclusion}
This paper presents STAR, Skeletal Token Alignment and Rearrangement for both human-robot and human-human interaction recognition. To the best of our knowledge, our proposed approach is the first to introduce multi-modal interaction alignment into skeleton-based interaction recognition. 
Specifically, STAR comprises three core components. First, a skeleton encoder leverages Entity Rearrangement and Interactive Spatiotemporal Tokens to capture fine-grained interdependencies. Second, Visual Interaction Encoding adopts a Focus on Interactions strategy to attend to spatiotemporal regions critical to interaction understanding. Finally, a contrastive alignment objective bridges skeleton and visual modalities, while a refinement head further improves prediction accuracy.
Experiments demonstrate that our STAR can achieve state-of-the-art performance across four interaction recognition datasets. 

Although our method assumes known actor types to ensure a controlled and comparable setting, extending it to more unconstrained environments is a valuable direction for future work. One potential approach is to incorporate clustering-based role assignment, where actor roles are inferred dynamically based on spatial-temporal movement patterns. Another promising direction is to adopt a multi-task learning framework that jointly performs interaction recognition and actor type classification. Such extensions would improve the generalizability of our approach to open-world environments, enabling more robust and flexible interaction understanding. Moreover, we plan to explore end-to-end grounding strategies that eliminate the reliance on external detectors, aiming to further improve robustness and integration of the visual encoding process.

\section*{Acknowledgments}
This work was supported by National Natural Science Foundation of China (No. 62473007), Natural Science Foundation of Guangdong Province (No. 2024A1515012089), Shenzhen Innovation in Science and Technology Foundation for The Excellent Youth Scholars (No. RCYX20231211090248064), and was in part by the Southern Marine Science and Engineering Guangdong Laboratory (Zhuhai) under Grant SML2024SP007.

\normalem
\bibliographystyle{IEEEtran}
\bibliography{Reference}

\end{document}